\pdfoutput=1
\documentclass{article}

\usepackage{microtype}
\usepackage{graphicx}
\usepackage{subfigure}
\usepackage{booktabs} 

\usepackage{multirow}
\usepackage[table,xcdraw]{xcolor}
\usepackage[normalem]{ulem}
\useunder{\uline}{\ul}{}

\usepackage[hyphens]{url}
\PassOptionsToPackage{hyphens}{url}\usepackage{hyperref}
\usepackage{tabularx}
\usepackage{makecell}

\usepackage[accepted]{icml2024}

\usepackage{amsmath}
\usepackage{amssymb}
\usepackage{mathtools}
\usepackage{amsthm}

\usepackage{calc} 

\usepackage[capitalize,noabbrev]{cleveref}

\theoremstyle{plain}

\theoremstyle{definition}

\theoremstyle{remark}

\usepackage[textsize=tiny]{todonotes}
\usepackage{longtable}

\icmltitlerunning{How well do LLMs cite relevant medical references? An evaluation framework and analyses}

\begin{document}

\twocolumn[
\icmltitle{How well do LLMs cite relevant medical references? An evaluation framework and analyses}




\icmlsetsymbol{equal}{*}

\begin{icmlauthorlist}
\icmlauthor{Kevin Wu}{equal,stanford}
\icmlauthor{Eric Wu}{equal,stanford}
\icmlauthor{Ally Casasola}{stanford}
\icmlauthor{Angela Zhang}{stanford}
\icmlauthor{Kevin Wei}{usc}
\icmlauthor{Teresa Nguyen}{stanford}
\icmlauthor{Sith Riantawan}{usc}
\icmlauthor{Patricia Shi}{loma}
\icmlauthor{Daniel Ho}{stanford}
\icmlauthor{James Zou}{stanford}
\end{icmlauthorlist}

\icmlaffiliation{stanford}{Stanford University}
\icmlaffiliation{usc}{University of Southern California}
\icmlaffiliation{loma}{Loma Linda University}

\icmlcorrespondingauthor{Kevin Wu}{kevinywu@stanford.edu}

\icmlkeywords{Machine Learning, ICML}

\vskip 0.3in
]



\printAffiliationsAndNotice{\icmlEqualContribution} 

\begin{abstract}
Large language models (LLMs) are currently being used to answer medical questions across a variety of clinical domains. Recent top-performing commercial LLMs, in particular, are also capable of citing sources to support their responses. In this paper, we ask: do the sources that LLMs generate actually support the claims that they make? To answer this, we propose three contributions. First, as expert medical annotations are an expensive and time-consuming bottleneck for scalable evaluation, we demonstrate that GPT-4 is highly accurate in validating source relevance, agreeing 88\% of the time with a panel of medical doctors. Second, we develop an end-to-end, automated pipeline called \textit{SourceCheckup} and use it to evaluate five top-performing LLMs on a dataset of 1200 generated questions, totaling over 40K pairs of statements and sources. Interestingly, we find that between $\sim$50\% to 90\% of LLM responses are not fully supported by the sources they provide. We also evaluate GPT-4 with retrieval augmented generation (RAG) and find that, even still, around 30\% of individual statements are unsupported, while nearly half of its responses are not fully supported. Third, we open-source our curated dataset of medical questions and expert annotations for future evaluations. Given the rapid pace of LLM development and the potential harms of incorrect or outdated medical information, it is crucial to also understand and quantify their capability to produce relevant, trustworthy medical references.

\end{abstract}

\section{Introduction}
\label{intro}
Large language models (LLMs) are increasingly considered for use in healthcare. Although no commercially available LLMs are currently approved by the FDA for use in medical decision support settings \cite{Gilbert2023-kq}, top-performing LLMs like GPT-4, Claude, and Med-PaLM have nonetheless demonstrated superior performance over clinicians on medical exams like the US Medical Licensing Exam (USMLE) \cite{Brin2023-im,Singhal2023-qu,Singhal2023-wi}. LLMs have already made their way into patient care today, from being used as chatbots for mental health therapy \cite{Ingram2023-yb,Maples2024-af} to users finding diagnoses for uncommon diseases that physicians missed \cite{Holohan2023-un}. A growing number of clinicians report using LLMs in their clinical practice or education \cite{Temsah2023-wb,Tangadulrat2023-cd}.

\begin{figure}[t!]
\includegraphics[width=1\columnwidth]{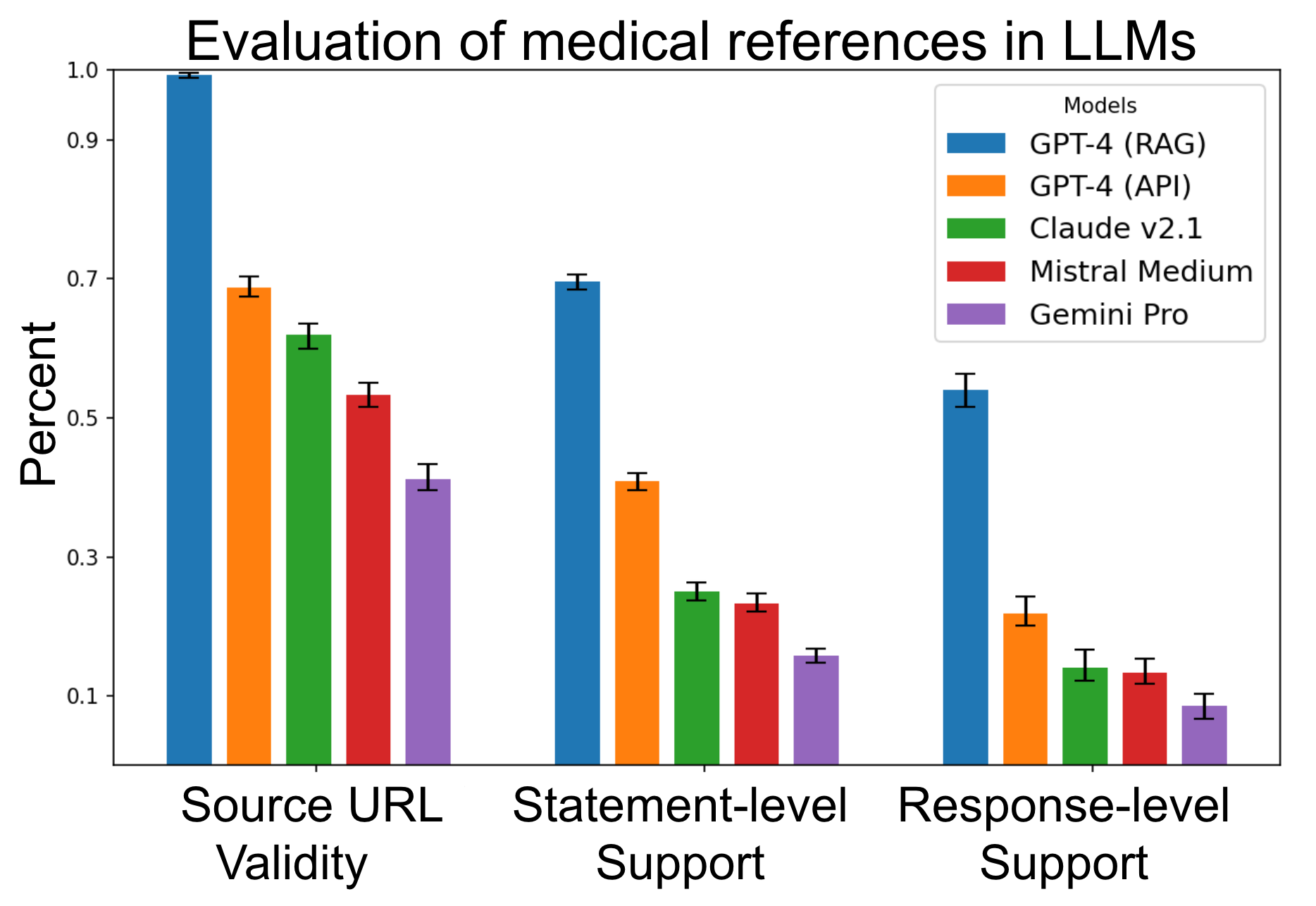}
\caption{Evaluation of the quality of source verification in LLMs on medical queries. Each model is evaluated on three metrics. \textit{Source URL Validity} measures the proportion of generated URLs that return a valid webpage. \textit{Statement-level Support} measures the percentage of statements that are supported by at least one source in the same response. \textit{Response-level Support} measures the percentage of responses that have all their statements supported. Full numerical results are displayed in Table \ref{tab:main_results}.}
\label{fig:main_results}
\end{figure}

\begin{figure*}[t!]
\vskip 0.2in

\includegraphics[width=1.0\textwidth]{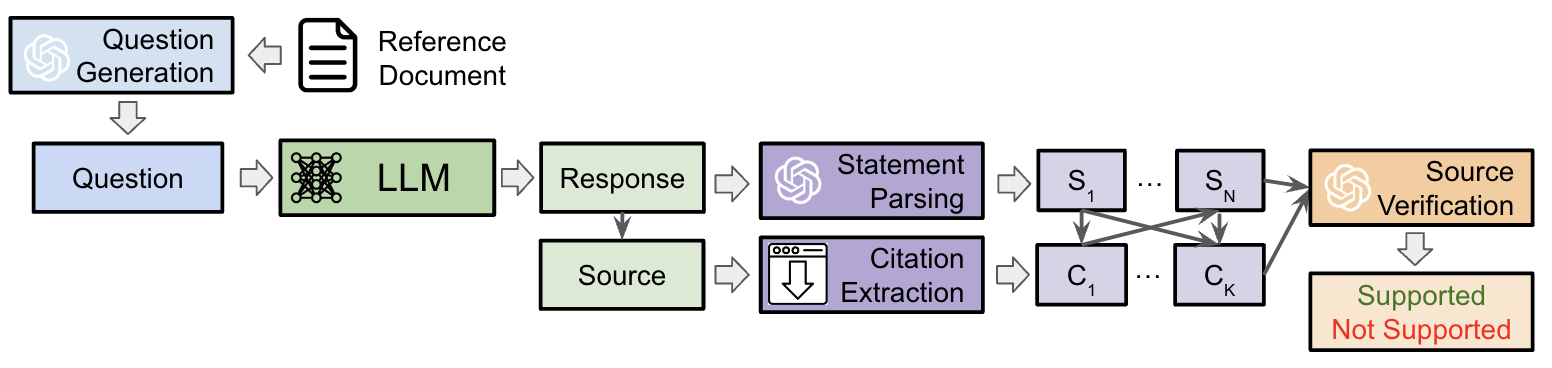}
\caption{Schematic of the \textit{SourceCheckup} evaluation pipeline. To start, GPT-4 generates a question based on a given medical reference text. Each evaluated LLM produces a response based on this question, which includes the response text along with any URL sources. The LLM response is parsed for individual medical statements, while the URL sources are downloaded. Finally, the Source Verification model is asked to determine whether a given medical statement is supported by the source text and to provide a reason for the decision.}
\label{fig:schematic}
\end{figure*}

However, LLMs are prone to hallucination, where the model generates statements not backed by any source \cite{Pal2023-up,Sun2024-mc,Ahmad2023-uc}. Particularly in the medical domain, this can erode user trust and potentially harm patients by providing erroneous advice \cite{Dash2023-il,Daws2020-ov} or discriminate based on patient backgrounds \cite{Nastasi2023-qk}. Lack of trust is commonly cited as the number one deterrent against clinicians adopting LLMs in their clinical practice \cite{Zawiah2023-nx,Abouammoh2023-bo}, and in particular, the inability of LLMs to generate supporting sources for medical statements in their responses \cite{Jansz2023-zw}.

\begin{figure*}[t!]
\vskip 0.2in
\begin{center}
\includegraphics[width=\textwidth]{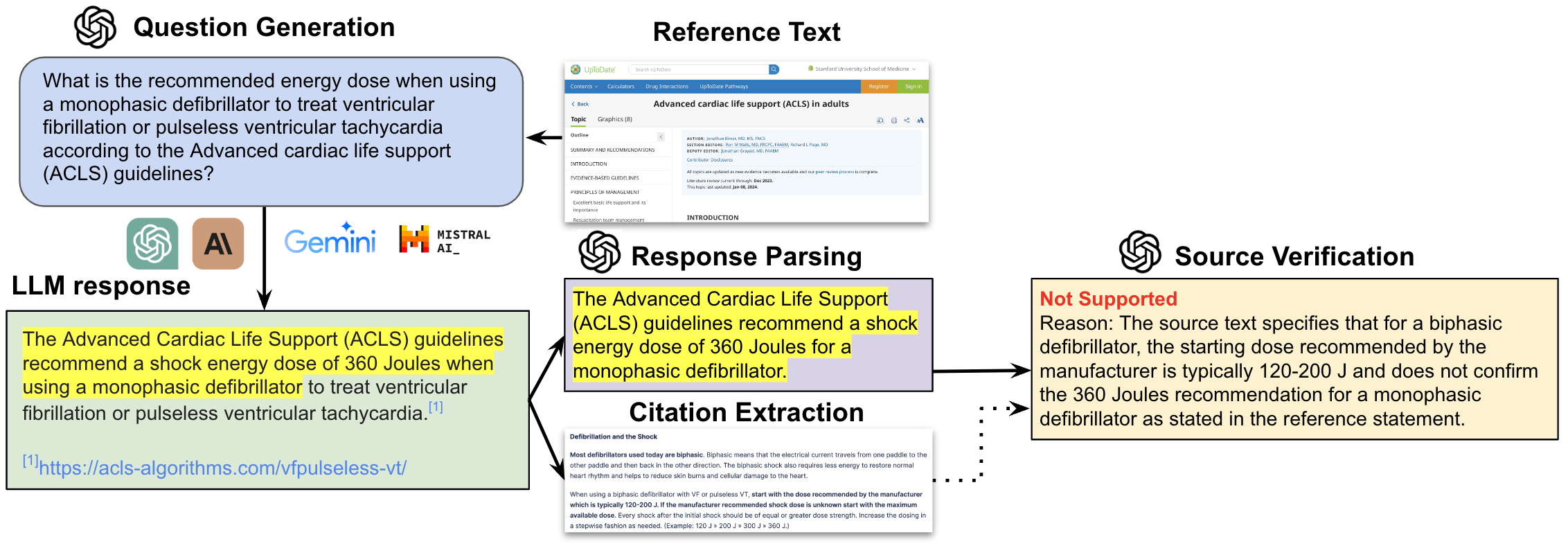}
\caption{An end-to-end example of the \textit{SourceCheckup} evaluation framework based on a real response from GPT-4 (RAG). A question is generated based on the contents of a medical reference text. The question is posed to an LLM, and the response is parsed into statements and sources. Each statement-source pair is automatically scored by the Source Verification model as supported (i.e. the source contains evidence to support the statement) or not supported. }
\end{center}
\label{fig:example}
\end{figure*}

The need to cite the sources for medical statements goes beyond gaining clinician and patient trust -- there is also an urgent regulatory case as well \cite{Hacker2023-zy}. The US Food and Drug Administration (FDA) has repeatedly called for regulating LLMs used as decision support tools \cite{Baumann2023-vd,Taylor2023-vr}.
Assessing the degree to which LLMs reliably convey existing, trustworthy medical knowledge is important for informing future regulatory frameworks regarding medical LLMs.

LLMs should be capable of reliably providing relevant sources to allow users and regulators to audit the reliability of their statements. Recent advancements in LLM capabilities (e.g. improved instruction fine-tuning) have enabled models to routinely provide sources upon request. Retrieval augmented generation (RAG), in particular, allows models to perform real-time searches for web references relevant to the query. However, even if the references are from valid and legitimate websites, it is still unclear the extent to which these provided sources contain content that actually supports the claims made in the model's generated responses. To this end, we make the \textbf{following contributions}. First, given the costly nature of high-quality medical expert annotations and the rapid pace of ongoing development in LLMs, we propose an automated evaluation framework, called \textit{SourceCheckup}, to create medical questions and to score how well LLMs can provide relevant sources in their answers to these questions. We verify that this framework is highly accurate, finding 88\% agreement with a consensus of three US-licensed medical physicians. Second, we evaluate top-performing, commercially available LLMs (GPT-4 (RAG and API), Claude v2.1, Mistral Medium, and Gemini Pro) and find that models without access to the Web only produce valid URLs between 40\% to 70\% of the time. Retrieval-augmented generation (RAG)-enabled GPT-4, which has search engine access, does not suffer from URL hallucination, but still fails to produce references that support all the statements in the response nearly half of the time. Finally, we open source our dataset of 1200 medical questions created from  webpages pulled from the Mayo Clinic, UpToDate, and Reddit, as well as a clinician-annotated subset of 284 question/answer pairs. Our findings highlight an important gap in the viability of LLMs for clinical medicine and have crucial implications for the medical adoption of LLMs.

\section{Related Works}

There is a growing body of work on measuring and improving source attribution in language models \cite{Li2023-pg,Rashkin2021-sm,Gao2023-df}.
Benchmark datasets introduced in works such as WebGPT \cite{Nakano2021-rr}, ExpertQA \cite{Malaviya2023-vl}, WebCPM \cite{noauthor_undated-iv}, and HAGRID \cite{Kamalloo2023-tw} aggregate open-domain subjects from web pages like Wikipedia in a question-answer format. However, the evaluations of these datasets were performed by manual human verification, which can be costly and time-intensive \cite{Chen2023-mn} and difficult to replicate.

There have been several works recently that have demonstrated the usefulness of using language models themselves in automatically scoring source attribution from LLMs. For instance, ALCE \cite{Gao2023-df}, AttributedQA \cite{Bohnet2022-yd}, and GopherCite \cite{Menick2022-oh} use supervised language models to perform automated evaluation of LLMs.
More relevantly, given the advent of powerful instruction-fine-tuned LLMs, FactScore \cite{Min2023-jp} and AttrScore \cite{Yue2023-tm} demonstrate that ChatGPT can be used as a useful evaluator of source attribution, but ChatGPT itself performs poorly when evaluated for source attribution \cite{Zuccon2023-jt, Liu2023-fe}.

Our proposed method makes three novel contributions. First, we construct the first dedicated corpus of medical-specific statement-source pairs (over 40K examples from over 1200 reference pages across three websites). Second, we provide evidence that GPT-4 is a highly effective evaluator of source attribution in the medical domain by showing strong agreement with a panel of three US-licensed medical doctors. Third, we use our automated framework to evaluate five state-of-the-art, commercially available LLMs commonly used by patients and clinicians today.

\begin{figure}[t]
\includegraphics[width=1.0\columnwidth]{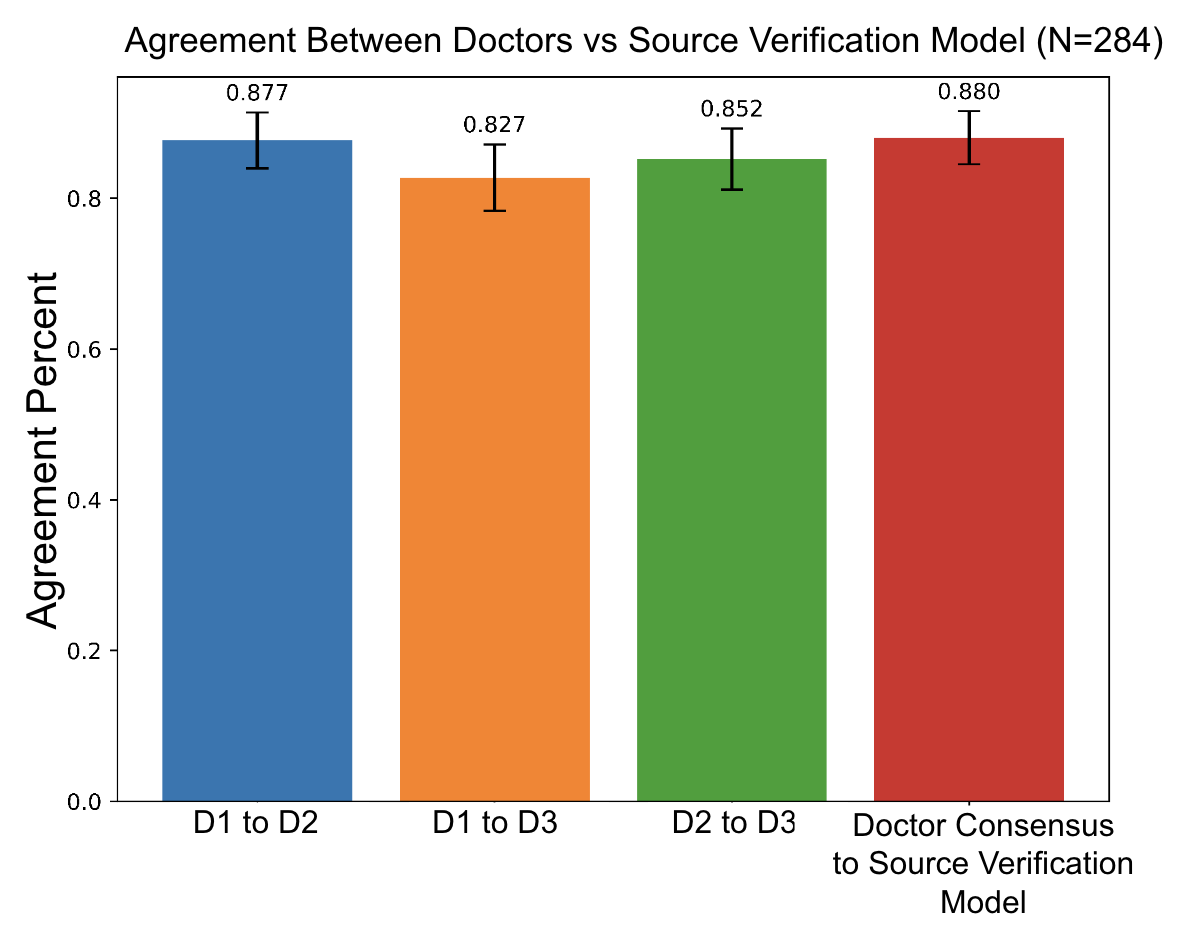}
\caption{Agreement between the Source Verification model and doctors on the task of source verification. We asked three medical doctors (D1, D2, and D3) to determine whether pairs of statements and source texts are supported or unsupported. We found that the Source Verification model has a higher agreement with the doctor consensus than the average agreement between doctors.}
\label{fig:agreement_figure}
\end{figure}

\section{Methods}

\subsection{LLM Evaluations}
Our analysis focuses on evaluating five top-performing LLMs: GPT-4 (RAG), GPT-4 (API), Claude v2.1 (API), Mistral Medium (API), and Gemini Pro (API). These models were chosen as they represent current leading LLMs \cite{Singhal2023-qu, Eriksen_Alexander_V2023-yw, Strong2023-vf, Tatsu_undated-bp} as of February 2024. We use \textit{gpt-4-1106-preview} as the GPT-4 API endpoint, and all model APIs were queried for responses on 1/20/24. GPT-4 (RAG) refers to GPT-4's web browsing capability powered by Bing. Other RAG models such as Perplexity.AI or Bard are currently unavailable for evaluation due to a lack of API access with sources, as well as restrictions on the ability to download their web results. For example, while \verb|pplx-70b-online| produces results with online access, it does not return the actual URLs used in those results. Gemini Pro is available as an API, but Bard's implementation of the model with RAG is unavailable via API.

\subsection{\textit{SourceCheckup} Evaluation Framework}
Our proposed pipeline consists of four modules: (1) Question Generation, (2) LLM Question Answering (3) Statement and URL Source Parsing, and (4) Source Verification. A schematic of this pipeline is found in Figure \ref{fig:schematic}, and an example is shown in Figure \ref{fig:example}. The prompts used for each of the following sections are detailed in Table \ref{tab:prompts}.

\subsubsection{Question Generation}
Given that medical question datasets such as PubMedQA \cite{Jin2020-kw} and MedQA \cite{Jin2019-of} consist of fixed question sets that are susceptible to memorization, we propose a question generation framework to create novel medical questions that reflect real-world clinical question/answering. A reference text was given to GPT-4 with a prompt to produce a question based on the content of the text. While there exist many possible medical reference texts to choose from, our analysis includes webpages from three sources: MayoClinic (\url{www.mayoclinic.org}), UpToDate (\url{www.uptodate.com}), Reddit r/AskDocs (\url{www.reddit.com/r/askdocs}). These three sources represent three types of questions -- MayoClinic provides patient-facing fact pages, UpToDate provides physician-facing articles with a deeper level of medical detail, and Reddit's r/AskDocs questions include in-the-wild questions that often do not have clearly defined answers. None of our reference documents were taken from private datasets containing protected health information. For each of these sources, we downloaded the text from 400 different topic pages, totaling 1200 initial reference documents. We then used GPT-4 to generate a question from each document. Finally, we posed each question to each of the five LLMs. We include several examples of questions in Table \ref{tab:medical_questions}.

\subsubsection{LLM Question Answering}
We queried each LLM to provide a short response to the question, along with a structured list of sources that support the response. The prompt used for querying LLMs can be found in  Table \ref{tab:prompts}. To gather responses from the GPT-4 (RAG) model with web browsing capabilities, we found that the standard prompt was unable to trigger the web search RAG capabilities, so we provided a modified version of the prompt that explicitly asks the model to use Bing Search. In a minority of cases, the model did not return a response or returned an incomplete response. In this event, we provided the LLM an additional try before considering the response invalid.

\subsubsection{Statement Parsing}
To break up the response into individually verifiable statements, we used GPT-4 to parse the LLM responses. For example, the response \textit{"The proportion of HFE C282Y homozygotes
with documented iron overload-related disease
is 28.4\% for men and 1.2\% for women"} is broken into \textit{["The proportion of HFE C282Y homozygotes
with documented iron overload-related disease
is 28.4\% for men", "The proportion of HFE C282Y homozygotes
with documented iron overload-related disease
is 1.2\% for women"]}. Certain responses did not return any parsed medical statements, largely due to the nature of the question asked. For example, the model response \textit{"Could you please provide the document or specify the details of the treatment options and durations for acute bacterial rhinosinusitis mentioned in it?"}, does not return any parsed medical statements. Full details of the number of parsed statements, along with source counts, are found in Table \ref{tab:sample_sizes}.
In general, we found that GPT-4 (API) and Claude v2.1 could consistently follow the instruction's JSON formatting, whereas other models had varying rates of success. In cases where the model fails to provide a structured source list, we extracted and removed all URLs using regular expression matching from the original text and treated them as the sources provided.

\subsubsection{URL Source Parsing}
For each URL source provided in the response, we downloaded the source content of the URL. We only kept websites that returned a 200 status code, meaning the content can be returned. A small percentage ($<$ 1\%) of cases also included websites that cannot be accessed through an automated request due to the website blocking access to certain user agents. We downloaded PDF documents locally before extracting their text using a PDF-to-text converter. After the source content was extracted, we applied a pattern-matching expression to strip code tags, leaving only the plain text. Finally, we excluded source contents that exceed the 128K maximum token length of GPT-4. This accounted for approximately 1\% of all downloaded URLs.

\subsubsection{Source Verification}
We considered a statement to be supported if it can be attributed to at least one source provided by the LLM. While not all sources are usually intended to support each statement, we found the task of determining each LLM's intended statement-source pairings to be difficult to determine. For example, some LLMs provided a footnote after each statement, whereas others provided a list of links at the end of a paragraph. As such, we opted to simply consider all pairs of statements and sources in our evaluation. Given a list of statements and sources, each possible pair was checked for whether the source contained the relevant information necessary to support the statement. For instance, given $M$ statements and $N$ sources, each of the $M$ statements was checked against each of the $N$ sources for a total of $M \times N$ pairs. 
For each pair, we prompted GPT-4 with the statement and source content and asked it to score the pair. If a statement was supported by at least one source, it was considered "supported"; otherwise it was considered "not supported". To disambiguate the use of GPT-4 for source verification as well as evaluation, we refer to this task as the "Source Verification model" and the evaluated model by the full name (i.e. GPT-4 (RAG) or GPT-4 (API)).

\citet{Rashkin2021-sm} proposes and formalizes a framework called \textit{Attributable to Identifiable Sources (AIS)}, which defines the AIS score of a given language model's response, $y$, supported by evidence $A$ as 1 if a human reviewer would agree that \textit{"$y$ is true, given $A$"} or 0 if not. \cite{Gao2023-df} extends this to measure the average sentence-level AIS score:
$$
Attr_{AIS}(y, A) = {avg}_{s\in y} AIS(s, A)
$$
which is the percentage of statements within a response that is fully supported by $A$. We extend these two definitions of statement-level and response-level AIS score to statement-level and response-level support below.

We report three metrics to evaluate each model's source verification capabilities:
\\

(1) \textbf{Source URL Validity}: \textit{Given all the source URLs produced by the model, what percent are valid?} We define a valid URL as one that produces a 200 status code when requested and returns valid text (non-empty response).

$$=
\left( \frac{\text{\# URLs with status code 200}}{\text{Total Number of URLs}} \right)
$$
\\

(2) \textbf{Statement-level support}: \textit{What percent of medically relevant statements produced by the model can be supported by \textbf{at least one} source?} For each statement parsed from the responses, we checked it against all sources produced by the model response. For example, if a response has three statements and two sources, then there will be a total of six pairs of statement-source pairs that will be checked. A statement was considered supported if at least one of those sources was found to contain supporting text. 
$$= \left( \frac{\text{Statements Supported by $\geq 1$ Source}}{\text{Total Number of Statements}} \right)$$ We note that this metric does not penalize LLM responses for producing many irrelevant sources. To this end, we also report the percent of URLs that are not used in supporting any statement, found in Table \ref{table:additional_stats}.
\\
(3) \textbf{Response-level support}: \textit{What percent of responses have all their statements supported?} For each response, we checked whether that response contained all supported statements.
$$= \left( \frac{\text{Responses w/ All Statements Supported}}{\text{Total Number of Responses}} \right)$$

\begin{table*}[t]
\centering
\begin{tabular}{lllll}
\toprule
\textbf{GPT-4 (RAG)} & \textbf{GPT-4 (API)} & \textbf{Claude v2.1} & \textbf{Mistral Medium} & \textbf{Gemini Pro} \\ \midrule
{\fontsize{8}{9.6}\selectfont mayoclinic.org (16\%)} & {\fontsize{8}{9.6}\selectfont ncbi.nlm.nih.gov (20\%)} & {\fontsize{8}{9.6}\selectfont ncbi.nlm.nih.gov (28\%)} & {\fontsize{8}{9.6}\selectfont mayoclinic.org (25\%)} & {\fontsize{8}{9.6}\selectfont mayoclinic.org (25\%)} \\ \hline
{\fontsize{8}{9.6}\selectfont ncbi.nlm.nih.gov (10\%)} & {\fontsize{8}{9.6}\selectfont mayoclinic.org (15\%)} & {\fontsize{8}{9.6}\selectfont mayoclinic.org (11\%)} & {\fontsize{8}{9.6}\selectfont ncbi.nlm.nih.gov (18\%)} & {\fontsize{8}{9.6}\selectfont ncbi.nlm.nih.gov (14\%)} \\ \hline
{\fontsize{8}{9.6}\selectfont clevelandclinic.org (9\%)} & {\fontsize{8}{9.6}\selectfont cdc.gov (7\%)} & {\fontsize{8}{9.6}\selectfont cdc.gov (5\%)} & {\fontsize{8}{9.6}\selectfont cdc.gov (5\%)} & {\fontsize{8}{9.6}\selectfont webmd.com (8\%)} \\ \hline
{\fontsize{8}{9.6}\selectfont drugs.com (3\%)} & {\fontsize{8}{9.6}\selectfont uptodate.com (5\%)} & {\fontsize{8}{9.6}\selectfont aafp.org (4\%)} & {\fontsize{8}{9.6}\selectfont medlineplus.gov (2\%)} & {\fontsize{8}{9.6}\selectfont hopkinsmedicine.org (8\%)} \\ \hline
{\fontsize{8}{9.6}\selectfont cdc.gov (2\%)} & {\fontsize{8}{9.6}\selectfont medlineplus.gov (4\%)} & {\fontsize{8}{9.6}\selectfont medicalnewstoday.com (3\%)} & {\fontsize{8}{9.6}\selectfont uptodate.com (2\%)} & {\fontsize{8}{9.6}\selectfont cdc.gov (7\%)} \\

\bottomrule
\end{tabular}

\caption{Top five cited websites by LLM. The domain names of each model's cited sources are extracted and ranked, with the top five displayed in the table above. Of note, the NIH (\texttt{ncbi.nlm.nih.gov}), MayoClinic (\texttt{www.mayoclinic.org}), and CDC (\texttt{www.cdc.gov}) are among the top-cited URLs all five models.}
\label{tab:top_domains_by_model}
\end{table*}

\subsection{Expert validation of GPT-4 Automated Tasks}

Each of the three GPT-4 automated tasks (Question Generation, Response Parsing, and Source Verification) was validated against the annotations of US-licensed practicing medical doctors. 

\subsubsection{Question Generation and Response Parser}
To validate the performance of GPT-4 on the task of generating questions from reference medical documents, we asked a medical doctor to spot-check 100 pairs of documents and questions for relevance and logical integrity. To validate GPT-4's performance on parsing medical statements from free-text responses, we also asked a medical doctor to analyze 330 statements from 72 questions to check (1) if all the statements are found within the original response, and (2) if any statements are missing from the list of parsed statements.

\subsubsection{Source Verification}
A subset of the statement and sources (N=284) was selected from the GPT-4 API model responses (based on 72 questions) sourced from the MayoClinic and r/AskDocs datasets. Three medical doctors independently scored whether the LLM-generated source verification decision correctly identified a statement as supported or not supported by the provided source. They also optionally provided a reason justifying their decision. We then calculated the majority consensus of the doctors and report the percent agreement among each doctor, the doctor consensus, and the LLM-generated decision. Additionally, we also include a set of 100 statement-source pairs generated by Claude v2.1 to test whether GPT-4 (API) is biased in favor of its own model responses.

\begin{table*}[h]
\centering
\begin{tabular}{llllll}
\toprule
 & \textbf{GPT-4 (RAG)} & \textbf{GPT-4 (API)} & \textbf{Claude v2.1} & \textbf{Mistral Medium} & \textbf{Gemini Pro} \\ \midrule
\% of valid URLs from domain\\ with paywall & 3.40\% & 7.07\% & 4.11\% & 4.36\% & 4.76\% \\  \hline
\% of invalid URLs with \\archived page (N=250) & 0.00\% & 3.50\% & 6.50\% & 0.50\% & 3.50\% \\  \hline
\% of URLs from United States & 85.09\% & 92.14\% & 93.00\% & 90.21\% & 91.84\% \\ \bottomrule
\end{tabular}
\caption{URL Statistics by LLM. Across all five models evaluated, we found low rates of URLs originating from sites with paywalls. We also found low rates of previously existing, but now defunct, pages, suggesting that the invalid links outputted by the LLMs were indeed hallucinated. Finally, we saw that the sources that the LLMs cite are predominantly from US-based websites.}
\label{tab:url_analysis_ai_models}
\end{table*}

\subsection{URL Analysis}
We computed several key statistics from the total set of URLs cited per model. First, we determined which domain names contain content that is hidden behind a paywall or subscription model. Second, of the URLs that were deemed invalid due to a 404 error or similar “page not found” response, we assessed how many URLs were previously valid but are now outdated. To approximate this, we used the \href{https://archive.org/help/wayback_api.php}{Internet Archive Wayback Machine API}, which stores archived URLs. Second, we reported the top five domain names cited by each model.
Finally, we analyzed the origin of URLs in two ways. First, we determined which domain names are US-based vs non-US-based by performing a \href{https://pypi.org/project/python-whois/}{whois lookup} and looking for a valid country. If no country is returned, we defaulted to count domains with TLDs contained in \texttt{\{.com, .org, .gov, .edu, .info, .net\}} as US-based, and non-US-based if not.

\section{Results}

\begin{figure*}[t!]
\centering
\includegraphics[width=0.9\textwidth]{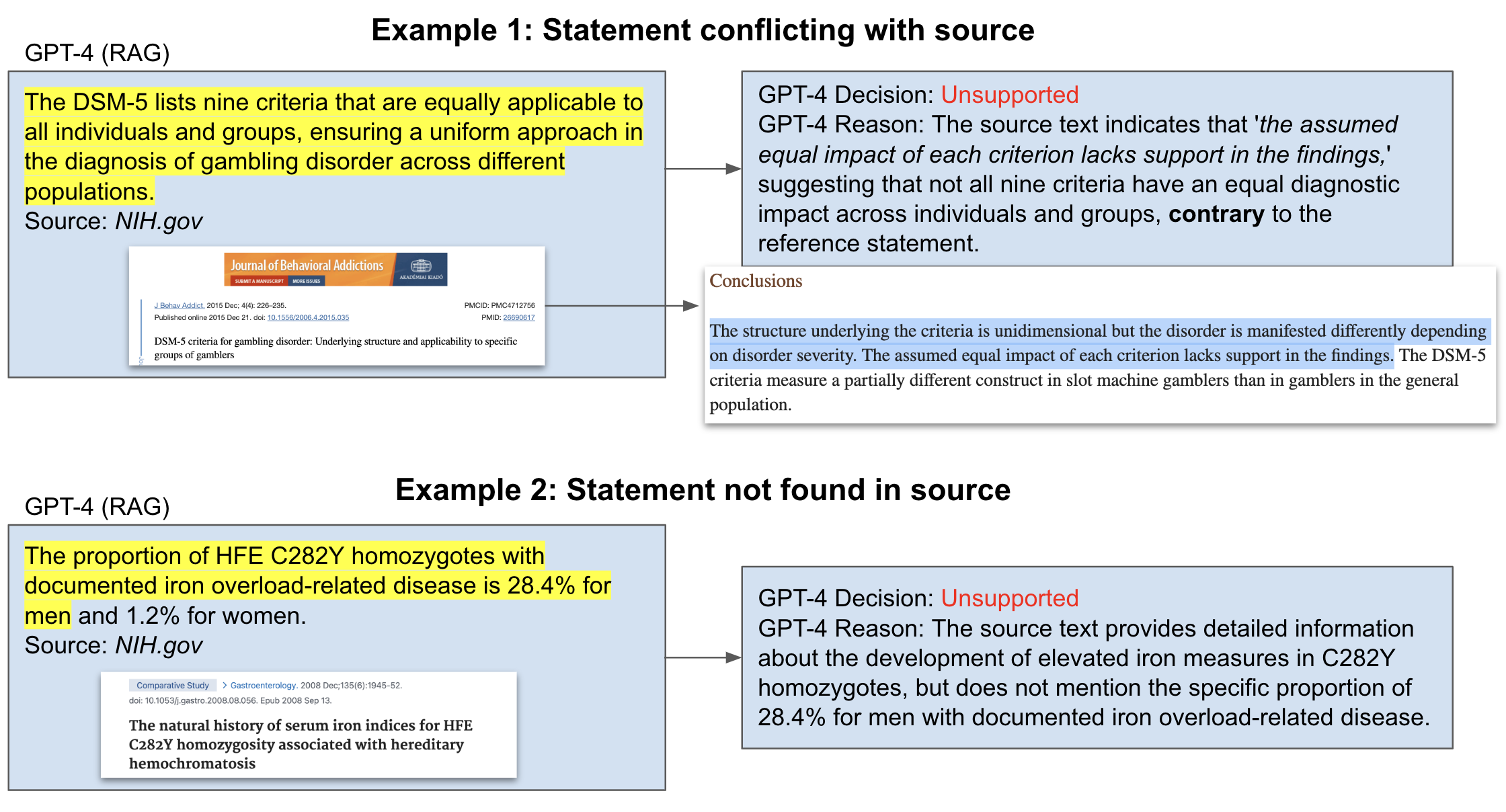}
\caption{Statements produced by GPT-4 (RAG) found to be unsupported by the Source Verification model. In the first example, the source provides information contrary to the statement in the response. In the second example, the statement is unsupported since it cannot be substantiated in the provided source.}
\label{fig:failure_modes}
\end{figure*}

\subsection{Expert validation of Automated Modules in SourceCheckup }
\subsubsection{Question Generation and Response Parsing}
A medical doctor verified that 100/100 of a random sample of generated questions are aligned with the reference document and can be answered. Additionally, 330/330 statements parsed from 72 responses were found to be contained in the original response, with 6 statements found to be in the original response but not contained within the parsed.
\subsubsection{Source Verification}
Our expert annotation of 284 statement-source pairings revealed that the Source Verification model performs as well as experts at determining whether a statement is supported by a source. We observed an 88.0\% agreement between the Source Verification model and the doctor consensus, and an 85.2\% average inter-doctor agreement rate (Figure \ref{fig:agreement_figure}, Table \ref{tab:agreement_matrix}, Figure \ref{fig:confusion_matrix}). We found no statistically significant difference between the doctor consensus annotations and Source Verification model annotations ($p=0.21$, unpaired sample t-test). We found 89\% agreement between the doctor's annotations and the Source Verification model's annotations on statement-source pairs generated from Claude v2.1. Compared to the 88\% agreement on statement-source pairs generated from GPT-4 (API), we find no statistically significance difference ($p=0.14$, unpaired sample t-test).

\subsection{Evaluation of Source Verification in LLMs}

Our full results of these three metrics across five models are found in Figure \ref{fig:main_results} and Table \ref{tab:main_results}. We found that GPT-4 (RAG) is the highest-performing model in terms of providing citation, mainly driven by its unique ability among the models to have access to the internet via search. However, we still found that its response-level support is only 54\%.  We provide examples of failures from GPT-4 (RAG) in Figure \ref{fig:failure_modes}, where one statement is not found due to it not being mentioned, and another is actually contradicted by a provided source. Additionally, the other four API-endpoint models all had much lower rates across the board. For example, GPT-4 (API), the currently best performing LLM \cite{Tatsu_undated-bp}, only produced valid URLs around 70\% of the time. On the other end, we found that Gemini Pro's API only produced fully supported responses about 7\% of the time.

As additional human expert validation, we randomly sampled 110 statement-source pairs produced by GPT-4 (RAG) that have been categorized by the Source Verification model as unsupported by any of the sources provided and had doctors assess each pair. The doctors agree with the Source Verifier 95.8\% (91.8\%-98.7\%) of the time. Among the 110 statement-source pairs provided by GPT-4 (RAG), the doctors confirmed that 105 statements are not supported by any source provided by GPT-4 (RAG). This result shows that retrieval augmentation by itself is not a silver bullet solution for making LLMs more factually accountable.

While the four API-endpoint models produced sources in $>$99\% of responses when prompted, we find that GPT-4 (RAG) fails to produce sources in over 20\% of responses, even when explicitly prompted to do so (Figure \ref{fig:breakdown}), partially contributing to its low response-level support.

\subsection{Breakdown by Question Source}
Next, we ask whether the type of question affects the quality of sources provided by LLMs. We found that the question source significantly affected every model's ability to produce supporting sources (detailed in Figure \ref{fig:by_source}). For example, while the response-level support for questions from MayoClinic is close to 80\% for GPT-4 (RAG), this drops precipitously to around 30\% on Reddit r/AskDocs ($p<0.001$ using unpaired sample t-test). Whereas questions from MayoClinic can be more directly answered from single sources, the Reddit r/AskDocs questions are more open-ended and often require pulling sources from a wide variety of domains. We found that questions from UpToDate was also more challenging for models to support compared to MayoClinic, as the questions produced from this source are typically more technical in nature.

\subsection{URL Analysis}
 We found that the URLs generated by the LLMs are predominantly from health information websites like mayoclinic.com or government health websites (e.g. nih.gov, cdc.gov) (Table \ref{tab:top_domains_by_model}).  We also found low rates of URLs coming from paywalled or defunct webpages (Table \ref{tab:url_analysis_ai_models}). Interestingly, most sources are from US-based websites (average of 90.46\%), with GPT-4 (RAG) having the highest proportion of non-US sources (14.9\%). Finally, we found that most sources are from .org or .gov domain names, indicating an origin of professional/non-profit organizations and governmental resources (Figure \ref{fig:sources_pie}).

\begin{table*}[t]
\centering
\begin{tabular}{l c c c}
\hline
\textbf{Model} & \textbf{\shortstack{Number of\\ Statements per Response}} & \textbf{\shortstack{Number of\\ Valid Sources per Response}} & \textbf{\shortstack{\\\% Valid URLs Not\\ Supporting Any Statement}} \\
\hline
GPT-4 (RAG) & 6.18 (6.02, 6.38) & 1.70 (1.62, 1.76) & 0.041 (0.033, 0.050) \\
GPT-4 (API) & 4.22 (4.13, 4.35) & 2.11 (2.07, 2.13) & 0.436 (0.413, 0.455) \\
Claude v2.1 & 3.84 (3.70, 3.94) & 1.88 (1.84, 1.91) & 0.646 (0.622, 0.669) \\
Mistral Medium & 3.29 (3.20, 3.37) & 2.18 (2.13, 2.23) & 0.647 (0.624, 0.672) \\
Gemini Pro & 2.58 (2.50, 2.68) & 1.71 (1.68, 1.75) & 0.678 (0.637, 0.713) \\
\hline

\hline
\end{tabular}
\caption{Statistics on responses and URLs from each of the five models. The first column reports the average number of sources in each response. The second column reports the average number of parsed statements in each response. The third column shows the percentage of URLs used in sources that do not support any of the statements within the response. Each cell contains the mean, along with the 95\% bootstrapped confidence interval (in parentheses).}
\label{table:additional_stats}
\end{table*}

\section{Discussion}

Sourcing high-quality medical annotations can be prohibitively costly and difficult to find. While previous works have used LLMs to confirm source attribution, our work is the first to validate automated medical source verification with a panel of medical experts. Our fully automated framework allows for the rapid development of novel question-answering datasets while reducing the need for additional manual annotations. This capability is key, especially in the field of clinical medicine, where standard-of-care and up-to-date knowledge is constantly evolving.

Our results highlight a significant gap in the current LLMs and the desired behavior in medical settings. Regulators, clinicians, and patients alike require that model responses be trustworthy and verifiable. Central to this is that they can provide reputable sources to back their medical claims.
Given that LLMs are predominantly trained on next-token prediction, it is unsurprising that models would provide hallucinated URLs or related, but incorrect, URLs as sources. To remedy this issue, models should be trained or fine-tuned directly to provide accurate source verification. RAG models show promise, as they can directly pull information from articles via search engines. However, we find that a substantial fraction of the references provided by RAG do not fully support the claims in GPT-4 (RAG)'s responses. This might be due to the LLM extrapolating the retrieved information with its pretraining knowledge or hallucination.  

An important distinction of our work is that we emphasize verifying whether the LLMs' generated statements are \textit{grounded} in verifiable sources, rather than directly assessing the \textit{correctness} of each claim. We take this approach because the nature of whether a claim is true or false can be up to subjective interpretation -- indeed, even medical experts may disagree over the degree to which a medical claim is fully factual. Instead, by directly verifying whether a claim is backed by an accessible source, we enable the user to verify and judge the veracity of the underlying source.

Under Section 230 of the Communications Decency Act, websites like Twitter or WebMD are not regulated by the US FDA, since they simply act as an intermediary for, rather than author of, medical information \cite{Haupt2023-uz}. However, it is unclear if this existing legal protection is likely to apply to LLMs since they can extrapolate and hallucinate new information. Additionally, the existing regulatory framework for AI software medical devices may also not apply to LLMs, as they do not have constrained, deterministic outputs \cite{Gottlieb2023-ot}. Thus, assessing the degree to which LLMs reliably convey existing, trustworthy medical knowledge is important for informing future regulatory frameworks regarding medical LLMs.

In our breakdown of source verification by question source, we find that models perform significantly worse on questions sourced from Reddit r/AskDocs versus either MayoClinic or UpToDate. This is significant, as questions from Reddit are user-generated whereas the other two sources are vetted by medical professionals and, in the case of UpToDate, intended for clinician use. One potential reason for this divergence is that user-generated questions tend to reflect a more diverse distribution of topics and more variable reading levels than medical reference sites, which tend to use precise medical terminology. Another hypothesis is that user-generated questions may contain erroneous premises for which LLMs have the propensity to affirm, known as contra-factual bias \cite{Dahl2024-jd}. In this same vein, we also found that the cited URLs are from US-based sources over 90\% of the time, which may potentially reflect American patient-centric medical evidence and standard of care. It is important thus for LLMs to adequately perform source verification to serve a wide range of users -- both laypersons and medical professionals, as well as sources that represent their intended demographic. 

Measuring source verification in models is also not intended to be considered in isolation. For example, one could trivially perform perfectly by quoting verbatim from a set of known sources (e.g. Google search). Instead, this benchmark should be used with other quality-based evaluation metrics to highlight inherent trade-offs in LLMs when they extrapolate information. To this end, we are releasing all of our curated data and expert annotations as a community resource.  

We believe that going forward, source verification is key to ensuring that doctors have accurate and up-to-date information to inform their clinical decision-making and providing a legal basis for LLMs to be used in clinic. Indeed, accurate source verification extends beyond the medical domain, and has apt applications in other fields like law (e.g. case law) and journalism (e.g. fact checking) as well.

\section{Impact Statement}
This paper presents work whose goal is to advance the field of Machine Learning, particularly in healthcare and trustworthy AI.

\bibliography{example_paper}
\bibliographystyle{icml2024}

\newpage
\appendix
\onecolumn
\section{Appendix}

\begin{table*}[h]
\centering
\begin{tabular}{lccc}
\hline
\textbf{Model} & \textbf{Source URL Validity} & \textbf{Statement-level Support} & \textbf{Response-level Support} \\ \hline
GPT-4 (RAG) & \textbf{0.992} (0.989, 0.995) & \textbf{0.694} (0.685, 0.703) & \textbf{0.543} (0.519, 0.569) \\
GPT-4 (API) & 0.692 (0.677, 0.706) & 0.422 (0.409, 0.434) & 0.227 (0.210, 0.246) \\
Claude v2.1 (API) & 0.621 (0.602, 0.642) & 0.252 (0.240, 0.263) & 0.137 (0.118, 0.157) \\
Mistral Medium (API) & 0.536 (0.519, 0.553) & 0.219 (0.209, 0.228) & 0.124 (0.107, 0.140) \\
Gemini Pro (API) & 0.413 (0.395, 0.427) & 0.143 (0.134, 0.157) & 0.076 (0.063, 0.090) \\ \hline
\end{tabular}
\caption{Performance of each LLM on source verification metrics across three medical website datasets as visualized in Figure \ref{fig:main_results}. Source URL Validity measures the proportion of generated URLs that return a valid webpage. Statement-level support measures the percentage of statements that are supported by at least one source in the same response. Response-level support measures the percentage of responses that have all their statements supported. Each number is reported alongside the 95\% bootstrapped confidence intervals (in parentheses). The top-performing model, GPT-4 (RAG), is displayed in bold.}
\label{tab:main_results}
\end{table*}

\begin{figure*}[h]
\vskip 0.2in
\begin{center}
\includegraphics[width=\textwidth]{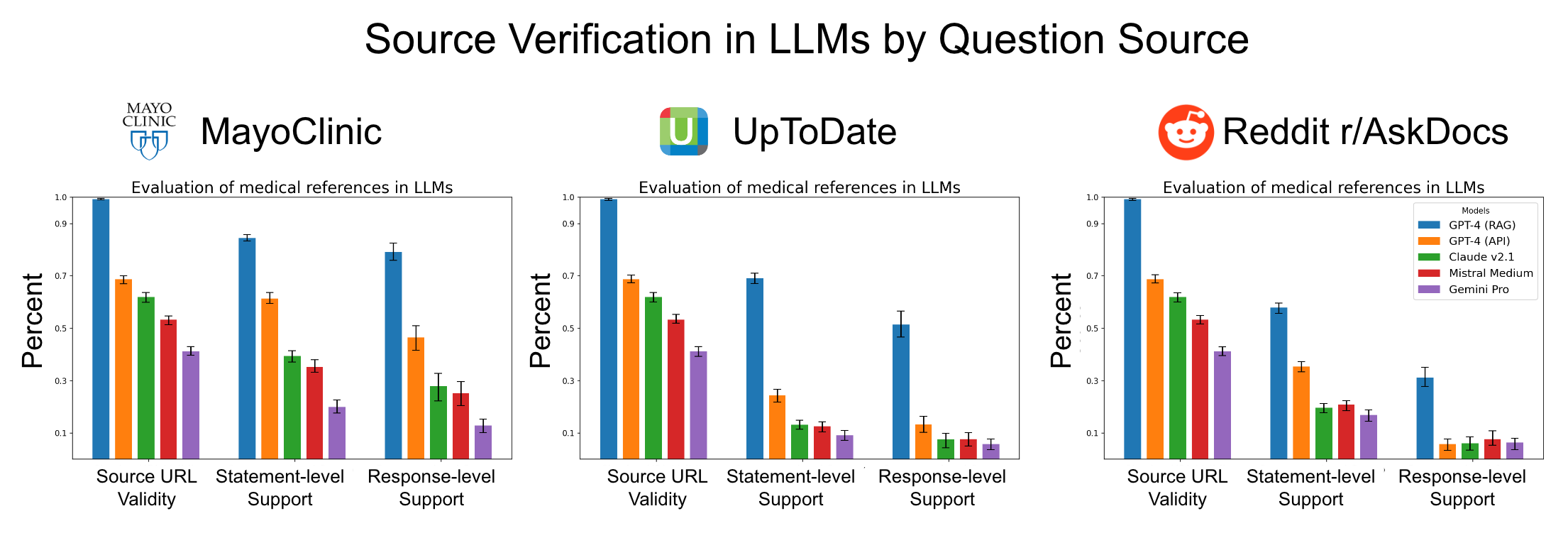}
\caption{Evaluation of source verification in LLMs, broken down by question source (dataset). In datasets with more statements per question, we found that models have a much harder time producing accurate sources at the statement and response level.}
\end{center}
\label{fig:by_source}
\end{figure*}

\begin{table*}[h]
\centering
\begin{tabular}{l|ccccc}
\hline
 & Doctor 1 & Doctor 2 & Doctor 3 & Consensus & Model \\
\hline
Doctor 1 & 1.00 & 0.88 (0.84, 0.92) & 0.83 (0.78, 0.87) & 0.93 (0.89, 0.95) & 0.85 (0.81, 0.89) \\
Doctor 2 & 0.88 (0.84, 0.92) & 1.00 & 0.85 (0.81, 0.89) & 0.95 (0.92, 0.98) & 0.88 (0.84, 0.92) \\
Doctor 3 & 0.83 (0.78, 0.87) & 0.85 (0.81, 0.89) & 1.00 & 0.90 (0.87, 0.94) & 0.81 (0.76, 0.86) \\
Consensus & 0.93 (0.89, 0.95) & 0.95 (0.92, 0.98) & 0.90 (0.86, 0.93) & 1.00 & 0.88 (0.85, 0.92) \\
Model & 0.85 (0.81, 0.89) & 0.88 (0.84, 0.92) & 0.81 (0.77, 0.86) & 0.88 (0.84, 0.92) & 1.00 \\

\hline
\end{tabular}
\caption{Agreement matrix between each doctor, the doctors' consensus decision, and the Source Verification model (referred to as "model") on source verification (N=284). We find high agreement between the doctor consensus and the Source Verification model (88\%). 95\% bootstrapped confidence intervals are included for each value in parentheses.}
\label{tab:agreement_matrix}
\end{table*}

\begin{table*}[h]
\centering
\begin{tabular}{l|cccc}
\hline
Model & \# Questions & \# Valid Responses& \# Parsed Statements & \# URLs \\ 
\hline
GPT-4 (RAG)     & 1200 & 1193 & 7408 & 2086 \\
GPT-4 (API)     & 1200 & 1198 & 5074 & 2928 \\
Claude v2.1     & 1200 & 1193 & 4608 & 2556 \\
Mistral Medium  & 1200 & 1150 & 3944 & 3004 \\
Gemini Pro      & 1200 & 1029 & 3102 & 2500 \\
\hline
\end{tabular}
\caption{In our analyses, each model is asked a total of 1200 questions. This table reports, for each model, how many questions received a valid response (where at least one parsable statement exists), how many parsed statements were produced from these valid responses, and how many sources were present in the valid responses.}
\label{tab:sample_sizes}
\end{table*}

\begin{figure}[h]
\includegraphics[width=\textwidth]{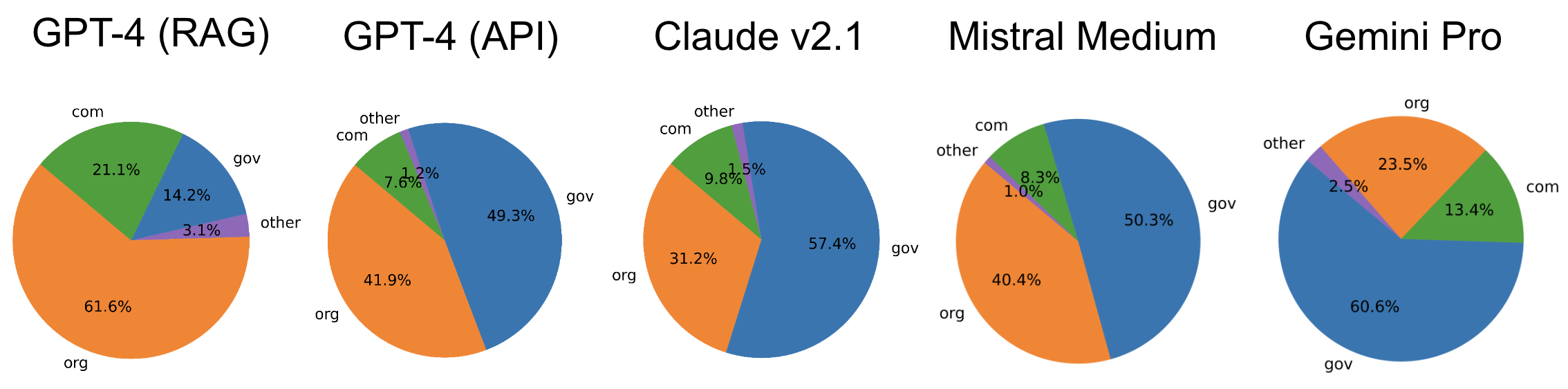}
\caption{Breakdown of the top-level domain (TLD) sources for each LLM. Each pie chart represents the proportion of how often each TLD appears in the total set of URL sources produced by each model.}
\label{fig:sources_pie}
\end{figure}

\begin{figure*}[b]
\centering
\includegraphics[width=0.6\textwidth]{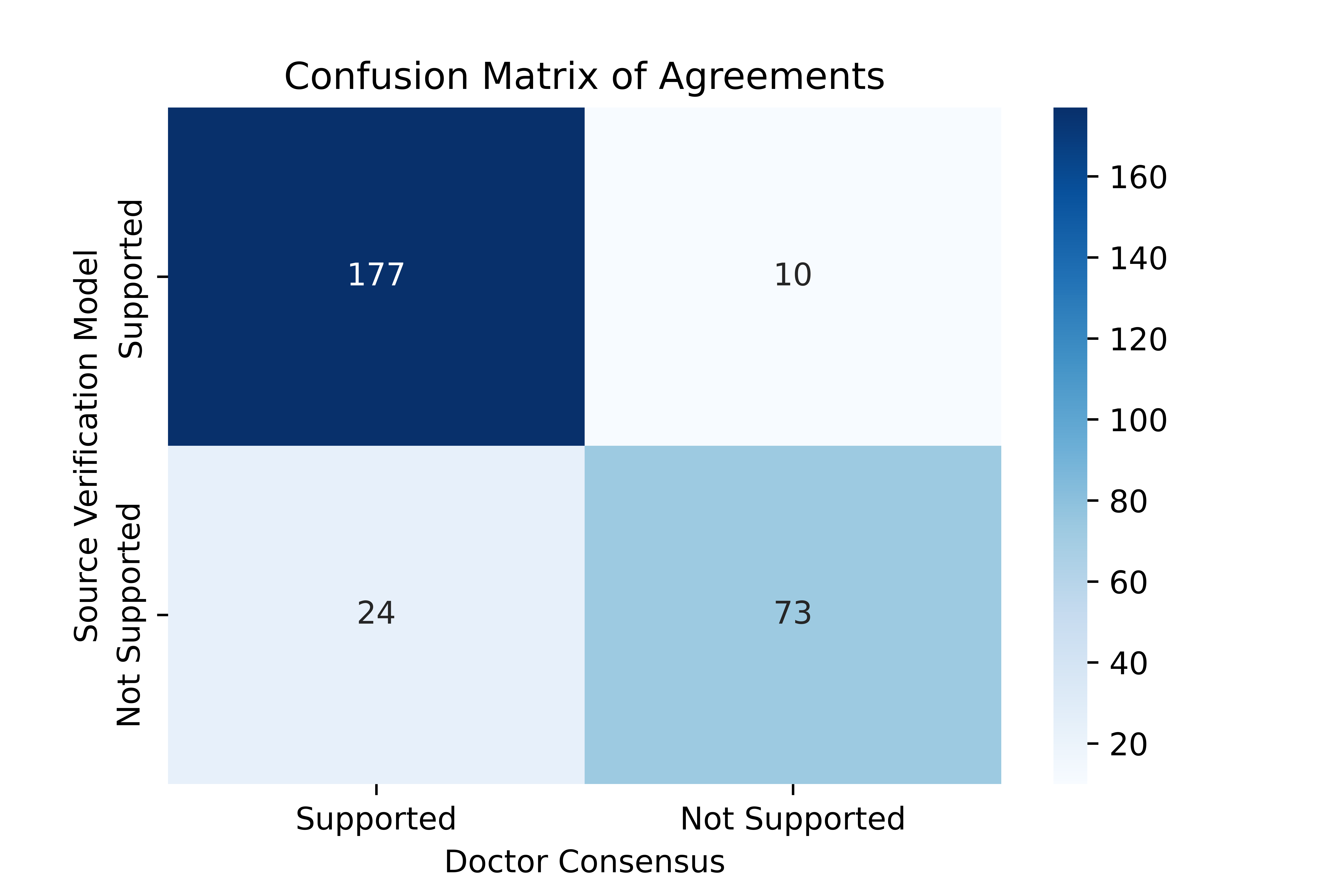}
\caption{Confusion matrix for human consensus vs Source Verification model agreement (N=284). Supported means that, for a given statement-source pair,  evidence was found in the source to support the statement. Not supported indicates that no such evidence exists.}
\label{fig:confusion_matrix}
\end{figure*}

\begin{figure*}[b]
\centering
\includegraphics[width=0.6\textwidth]{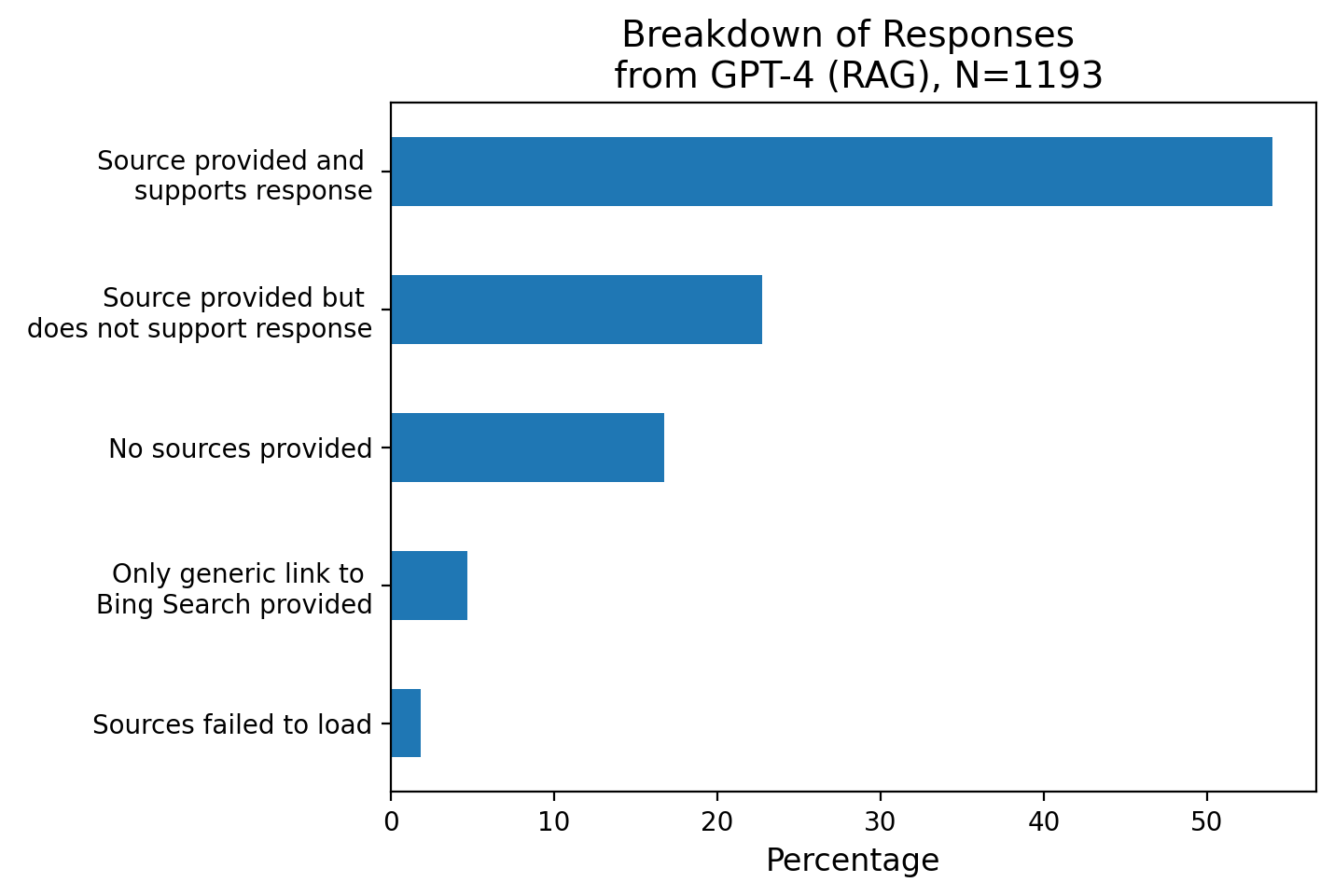}
\caption{A breakdown of the 1193 unsupported responses by GPT-4 (RAG). We find that nearly 17\% of the cases where a response is unsupported are due to a failure of the model to produce sources at all. Additionally, around 5\% of unsupported responses only have a link to a Bing Search as the source. Finally, around 2\% of cases were due to a failure of the GPT-4 (RAG) web interface to render a source successfully. Other (non-RAG) models are included as we find they all have $>99$\% of responses have at least one source provided.}
\label{fig:breakdown}
\end{figure*}

\begin{table}[h]
\centering
\begin{tabular}{@{}p{0.4\textwidth}p{0.1\textwidth}p{0.4\textwidth}@{}}
\toprule
\textbf{Question} & \textbf{Source} & \textbf{URL} \\ 
\midrule
What symptoms can suggest that a person's carpal tunnel syndrome may be severe enough to consider surgery as a form of treatment? & MayoClinic & {\scriptsize \url{https://www.mayoclinic.org/diseases-conditions/carpal-tunnel-syndrome/symptoms-causes/syc-20355603}} \\ 
What is the preferred antiparasitic treatment regimen for infants with confirmed or highly likely congenital toxoplasmosis? & UpToDate & {\scriptsize \url{https://www.uptodate.com/contents/congenital-toxoplasmosis-treatment-outcome-and-prevention}} \\
(19 yo F, 5'5, 170 pounds, medications are pristiq and klonopin) for the last 2 going in 3 days I have a constant pain under my right ribcage that's driving me insane. It's not severe but hurts when I press or lay on it. I have no other symptoms and google is not helping with my anxiety. I'm a but of a hypochondriac and don't know if I need to let it keep worrying me and see a dr or if it sound like nothing and I need to let my head stop getting the best of me. & Reddit & {\scriptsize \url{https://www.reddit.com/r/AskDocs/comments/18r0cyg/gallbladder_or_pulled_muscle/}} \\ \bottomrule
\end{tabular}
\caption{Examples of generated medical questions for each website. Each URL is downloaded and used to prompt GPT-4 to generate a question based on the content of the webpage. One representative example is shown from each of the three sources used in our dataset.}
\label{tab:medical_questions}
\end{table}

\begin{table}[h]
\centering
\begin{tabular}{|l|p{10cm}|}
\hline
\textbf{Role} & \textbf{Description} \\ \hline
Question Generation & You are given a prompt that contains medical information. Generate a question that tests a medical expert's understanding of the information contained within the prompt. Each question should provide sufficient context so that the question does not require the prompt to be fully understood. Do not reference the word 'prompt' in the question. The question should have a straightforward answer that can be easily verified with the prompt. In addition to the question, provide the answer that is succinctly taken from the prompt. Return a JSON where the key 'question' provides the question in a string and the key 'answer' provides the answer in a string. \\ \hline
Statement Parsing & Divide the following statement into distinct medical facts as a list in a JSON with the key '0'. Ignore statements that are not inherently medical facts, but rather express the limitations of the response. Each fact should have the full context of the full statement and retain the subject so that it is readable on its own without the context of the other facts. \\ \hline
Question Answering (API Models) & Answer the following medical question accurately and precisely in 2-3 sentences max. Provide URL sources to credible sources to substantiate every medical fact stated in your answer. Return a JSON where the key 'response' has the value of the text of the answer and the field 'sources' has the value of a list of source URLs. \\ \hline
Question Answering (GPT-4 RAG) & Answer the following medical question accurately and precisely in 2-3 sentences max and cite your sources with URLs. Use Bing Search. \\ \hline
Source Verification & Your task is to verify the accuracy of a reference statement against a provided source text. Based on your analysis, determine if the reference statement is confirmed by the source text or not. Respond in JSON format, where the evaluation score ("found" or "not\_found") is the key, and your reasoning is the value. \\ \hline
\end{tabular}
\caption{Prompts used for each component of the \textit{SourceCheckup} evaluation framework. Full details of where each prompt is used are described in the Methods section.}
\label{tab:prompts}
\end{table}


\end{document}